\pdfoutput=1
\documentclass[11pt]{article}

\usepackage[final]{ACL2023}

\usepackage{times}
\usepackage{latexsym}

\usepackage[T1]{fontenc}
\usepackage[utf8]{inputenc}

\usepackage{microtype}

\usepackage{inconsolata}

\usepackage{amssymb}
\usepackage{hyperref}
\usepackage{url}
\usepackage{graphicx}
\usepackage{booktabs} % To thicken table lines
\usepackage{blindtext}
\usepackage{amsthm}
\usepackage{soul}
\usepackage{float}
\usepackage{cleveref}

\title{Not All Similarities Are Created Equal: \\ 
Leveraging Data-Driven Biases to Inform GenAI Copyright Disputes}

\author{
Uri Hacohen$^{\lambda}\thanks{\;\;~~Equal.}$ \hspace{0.5cm} Adi Haviv$^{\tau*}$ \hspace{0.5cm}  Shahar Sarfaty$^{\tau}$ \hspace{0.5cm} \textbf{Bruria Friedman}$^{\lambda}$  \\\textbf{Niva Elkin-Koren}$^{\lambda}$ \hspace{0.5cm}\textbf{Roi Livni}$^{\eta}$ \hspace{0.5cm} \textbf{Amit H Bermano}$^{\tau}$ \\ \\
$^{\tau}$ School of Computer Science\hspace{0.04cm},
$^{\lambda}$ Faculty of Law \hspace{0.04cm},
$^{\eta}$ School of Electrical Engineering \hspace{0.04cm} \\
Tel Aviv University \\
\small{\texttt{adi.haviv@cs.tau.ac.il}}
}

\begin{document}

\maketitle

\begin{abstract}
The advent of Generative Artificial Intelligence (GenAI) models, including GitHub Copilot, Open AI GPT, and Stable Diffusion, has revolutionized content creation, enabling non-professionals to produce high-quality content across various domains. This transformative technology has led to a surge of synthetic content and sparked legal disputes over copyright infringement. To address these challenges, this paper introduces a novel approach that leverages the learning capacity of GenAI models for copyright legal analysis, demonstrated with GPT2 and Stable Diffusion models. 
Copyright law distinguishes between original expressions and generic ones (Scènes à faire), protecting the former and permitting reproduction of the latter. However, this distinction has historically been challenging to make consistently, leading to over-protection of copyrighted works. GenAI offers an unprecedented opportunity to enhance this legal analysis by revealing shared patterns in preexisting works. We propose a data-driven approach to identify the genericity of works created by GenAI, employing "data-driven bias" to assess the genericity of expressive compositions. This approach aids in copyright scope determination by utilizing the capabilities of GenAI to identify and prioritize expressive elements and rank them according to their frequency in the model's dataset.
The potential implications of measuring expressive genericity for copyright law are profound. Such scoring could assist courts in determining copyright scope during litigation, inform the registration practices at the Copyright Offices, allow registration of only highly original synthetic works, and help copyright owners signal the value of their works and facilitate fairer licensing deals. More generally, this approach offers valuable insights to policymakers grappling with adapting copyright law to the challenges posed by the era of GenAI.

\end{abstract}
\section{Introduction}
\label{sec:intro}

The emergence of Generative Artificial Intelligence (GenAI) models is fostering a radical transformation in the creative domain. By using GenAI models, non-professional users can now generate high-quality content such as text, images, music, or code. At the same time, however, GenAI is also disrupting copyright law. By reducing the costs and talent barriers associated with creative production, GenAI has flooded markets with a massive yield of syntactic content, leaving scholars and copyright regulators to ponder whether to merit copyright protection to such works \cite{Burk2023, Grimmelmann2016}. The approach proposed in this paper can help the stakeholders grappling with this question by informing them of the scope of unprotected elements in copyrighted works. In addition, numerous lawsuits are emerging, claiming that GenAI models are “21st-century collage tools” that infringe the copyrighted works used to train these models \citet{Andersen2023,Doe2022, Dogan2005}. These pending class action lawsuits will require courts to decide, first, whether unauthorized use of copyrighted works to train GenAI models constitutes copyright infringement and second, when does the output generated by GenAI models infringes the copyright of works included in the training dataset. Our paper only addresses the latter question. 

These challenges put increasing pressure on the legal system to discern which portions of copyrighted works are protected from unauthorized use and which portions are not protected and are, therefore, available for others to use. Copyright law’s ultimate goal is to foster the creation and dissemination of expressive works by granting authors only limited rights to their respective expressions \cite{leval1990}. That is because creation never occurs in a vacuum, and the author always draws upon preexisting materials by adapting them into an original expression. Consequently, some elements of copyrighted works that are building blocks of further creation, such as style, ideas, functions, and methods, are not protected against unauthorized uses \cite{elkin2023can}. In fact, copyright law encourages users to draw upon these unprotected materials to create new original works of authorship \cite{litman1990}.
Copyright law’s greatest challenge is in allocating legal entitlements to expressive works by consecutive authors. This task has always been a challenge because authorship derives from cultural contexts. Authors routinely engage with preexisting materials to convey a meaningful message. This task grows far more complicated when GenAI augments human creativity \cite{grimmelmann2009ethical}. Generative models draw upon an infinite number of preexisting works created by an often-unidentifiable multitude of authors. Consequently, allocating legal entitlements among all possible claimants becomes an insurmountable objective.
For decades, courts (and, to a lesser extent, the Copyright Office) used the concept of originality and related legal doctrines to calibrate the scope of legal protection for copyrighted works \cite{miller2009hoisting, parchomovsky2009originality}. In this article, we argue that GenAI may help the courts in making such calibrations. Copyright law requires works to be minimally original to be protectable, but the breadth of their originality will dictate the effective scope of these works’ protection against allegedly infringing uses. In copyright infringement litigation, when copyright owners ask the courts to enforce their rights against alleged infringers, courts implicitly assess the originality of copyrighted works and delineate their legal protections accordingly \cite{parchomovsky2009originality}. So far, courts have performed this task on an ad-hoc basis by applying numerous legal doctrines such as the idea/expression dichotomy, merger, Scènes à faire, substantial similarity, and fair use \cite{sas2014institute}. These doctrines are notoriously vague and unpredictable and, in practice, often lead to the over protection of preexisting works \cite{gibson2007, litman2008, tehranian2007,samuelson1996,vaidhyanathan2003}.  
GenAI introduces new opportunities to infuse legal analysis with quantitative measures. Since GenAI documents the output of human creativity on an unprecedented scale, it facilitates a systematic study of the concealed interconnections among elements of expressive works. Consequently, GenAI may facilitate the development of new and more accurate measures to identify expressions that have become generic and better appraise the originality of these works.
In this article, we argue that researchers could leverage the data-driven bias of GenAI machines to inform copyright infringement disputes and calibrate the scope of copyright protection. “Data-driven bias” is a fundamental characteristic of machine learning and affects the way in which GenAI models generalize \cite{10.1145/3528233.3530698}. The more commonly expressive features appear in the GenAI models’ training dataset (the more “generic” they are), the more likely GenAI models are to utilize them when generating new works. 
If the models’ datasets reflect the actual cultural state of specific expressive domains, the models’ data-driven bias could indicate the scope of legal protection that the law affords to copyrighted works. Copyright protection does not extend to generic elements that are not considered original, which, in turn, depends on their prevalence and cultural embedment. The more ubiquitous the compositions of elements are, and the more they are absorbed in preexisting works, the less likely they are to be considered original under copyright law. Thus, the law will consider common expressive elements unoriginal (narrowly protected), and uncommon expressive elements original (broadly protected). 
By harnessing data-driven bias to identify genericity and assess copyright originality, computer science research can inform the legal profession across various practical applications. For example, courts could use genericity findings to delineate the scope of copyright protection in copyright infringement analysis. Copyright Offices could use these assessments to decide whether to register (and afford protection) GenAI works, and copyright owners could use this assessment when negotiating copyright licenses. In this paper, we define and describe data-driven bias and explore its implications for copyright analysis. We also demonstrate why current proposals for applying technical procedures are misguided and less informative to the legal profession. We further highlight several ways by which the framework and methodology we propose could facilitate follow-on research to measure and score copyright originality.

\section{Related Work}
A growing number of researchers in recent years explore how to address legal problems by applying theories and methods of computer science. This literature seeks to narrow the gap between the vague and abstract concepts used by law by applying mathematical models to offer more rigor, coherent and scalable definitions into issues such as privacy \cite{dwork2018privacy}, or fairness and discrimination \cite{dwork2012fairness}. The study of copyright by computer science methods has only emerged recently.  \citet{scheffler2022formalizing}, for instance, proposed a framework to test substantial similarity by comparing Kolmogorov-Levin complexity with and without access to the original copyright work. 

In the context of generative models \citet{carlini2023extracting,haim2022reconstructing},  explore whether generative diffusion models memorize protected works that appeared in the models’ training set. This can be considered as a preliminary issue to the problem of establishing copyright infringement. However, as we discuss in \cref{sec:discussion}, memorization of input content does not necessarily equate with copyright infringement.
There is also active and thought-provoking discussion on how machine learning technologies are reshaping our understanding of copyright within the realm of law. \citet{asay2020independent} explores the question of whether AI system outputs should be subject to copyright. Additionally, \citet{grimmelmann2015copyright,lemley2020fair} explore the implications of copyright law for literary machines that extract content and manage databases of information.
Other works, like \citet{bousquet2020synthetic, vyas2023provable} attempt to evaluate copyright infringement in GenAI models using privacy-like notions. The contribution of these works lies in proposing a procedure to inform copyright analysis in line with the thesis of this paper. However, this approach may not scale easily, and it also falls short of providing nuanced information regarding the level of genericity, which could be crucial for resolving copyright legal disputes \cite{elkin2023can}.  
Another body of related work attempts to assign attribution scores to training content to understand the contribution of individual (or group) examples to model predictions \cite{koh2017understanding,ghorbani2019data,jia2019towards,pezeshkpour-etal-2021-empirical,osti_10348871}. These approaches tend to adopt techniques such as leave-one-out retraining or influence functions to understand model behavior, fix mislabeled examples, and debug model errors \cite{koh2017understanding}. Instant attrition approaches are useful to determine the source of generated outputs which can inform copyright infringement analysis or assign credit to contributors \cite{henderson2023foundation}. However, instant attribution approaches do not apprise the added value of specific outputs relative to the aggregated knowledge of GenAI models.  
Most similar to our approach are previous attempts by computer scientists to harness computational measurements for assessing creativity \cite{franceschelli2021creativity}. For instance, \citet{franceschelli2022deepcreativity} proposed using generative learning techniques to assess creativity based on Margaret Boden’s definition of value, novelty, and surprise \cite{boden2007creativity}. They also created a tool that executed this assessment called \textit{DeepCreativity}. However, these systems that rely on philosophical and psychological perspectives do not necessarily align with the legal principles of copyright law. 
%missing citation . No 29.

\section{Research Approach}%RESEARCH APPROACH
\subsection{Copyright Objectives}
Copyright law seeks “To promote the Progress of Science and useful Arts, by securing for limited Times to Authors and Inventors the exclusive Right to their respective Writings and Discoveries” \cite{usconstitution}. Accordingly, copyright law incentives creation and dissemination of original works by granting authors exclusive rights to their respective works \cite{mazervstein1954}. These rights ensure that authors can commercially exploit their works and sustain incentives to invest in creating future works. However, promoting progress is inconsistent with granting unlimited rights to control copyrighted materials. Instead, it often requires setting substantial limits on the rights granted to authors \cite{dirobilant2022property, cohen1998lochner, lemley2005property}. That is because creative processes are situated in cultural contexts which involve building upon existing works and ongoing interactions with preexisting materials.

For this reason, copyright protection extends only to the aspects of expressive works that the law considers original \cite{usc1990,feist1991,harper1985}. Originality requires that expressive subject matter originates from its authors rather than copied from someone else. Therefore, facts and discoveries are never eligible for copyright protection \cite{feist1991}. In addition, originality requires that expressive subject matter contains a modicum of creativity. Therefore, if a work simply reflects a set of widely used cultural elements, it cannot be reasonably attributed to any particular author. 

The legal standard of creativity is vague. Courts defined creativity only in a negative manner to include expressions that are not an “age-old practice, firmly rooted in tradition and so commonplace that it has come to be expected as a matter of course.” For example, in the seminal case of Feist v. Rural, the court held that an alphabetical arrangement of telephone subscribers in a white page directory is insufficiently creative to merit copyright protection, because it was a “time-honored tradition [that] does not possess the minimal creative spark required by the Copyright Act and the Constitution.”

%todo: add citations 35 till 38. they appeared messed up in the word. 

Expressive works do not need to meet a high threshold of originality to be eligible for copyright protection \cite{feist1991,goldstein2005copyright}. All expressive works originating with their authors that are fixed in a tangible medium and has a modicum of creativity are eligible. Nevertheless, the actual scope of legal protection that the law affords expressive works varies according to the degree of originality these works embed \cite{hacohen2024copyright}.

Courts apprise the originality of expressive works and delineate their copyright protection in the course of infringement litigation. When copyright owners ask the courts to enforce their rights against alleged infringers, courts must decide whether the alleged infringers’ unauthorized uses of copyrighted works constitute copyright infringement. When making such determinations, courts implicitly consider the originality level of the copyrighted
works \cite{miller2009hoisting, parchomovsky2009originality}. Specifically, courts engage in “analytic dissection,” based on a framework established in \textit{Computer Associations International, Inc. v. Altai, Inc.} \cite{computerassociations1992}. 

The Altai framework requires the courts to separate the copyrighted works into different levels of abstraction, filter out the non-original material, and then evaluate the similarity of the remaining original matter of the copyrighted work against the allegedly infringing works \cite{lemley2010}. 

Naturally, the smaller the portion of the original material within copyrighted works (and the less original that portion is), the less likely the courts to find unauthorized uses of such works to be infringing \cite{lemley1995}. Thus, unlike highly original works, which enjoy broad legal protection against different forms of derivative uses, limitedly original works will enjoy only narrow legal protection limited to cases of extreme verbatim copying \cite{AppleComputerVMicrosoftCorp}. 

In practice, both the filtration and the comparison stages of the Altai framework require the courts to make ad-hoc assessments of notoriously vague legal doctrines \cite{litman1990}. During filtration, the courts must assess which elements of the work are non-original (“generic”) because they have merged with an “idea” or been used long enough to become indispensable \cite{Samuelson2016}. Courts invoke numerous overlapping doctrines such as “merger,” “useful article,” and “scènes à faire” to make these assessments \cite{sas2014institute}.  For example, in  Acuff-Rose v. Jostens \cite{AcuffRoseVJostens}, the Court ruled that the defendant was free to copy the phrase “You’ve Got to Stand for Something” from the plaintiff’s song because the phrase was overused to become a “cliché.” Similarly, in Lotus v. Borland \cite{LotusVBorland}, the Court denied protection to the menu command hierarchy of the spreadsheet program Lotus 1-2-3. As Justice Boudin explained, the extensive use of these commends have made them into a “method of operation,” which is unqualified for copyright protection. 

Once the courts finish filtering out unprotected material, they move to the next stage of the Altai framework: comparing the similarity of the remaining material to the allegedly infringing work. If the courts find that the two are “substantially similar,” they will find for infringement unless they decide that the challenged use is qualified as “fair use” \cite{lemley2010}.  When the courts evaluate “substantial similarity” and “fair use,” they also consider the originality of the allegedly infringed copyrighted works \cite{lemley1995, AppleComputerVMicrosoftCorp,USC2012}. The less original the copyrighted works, the higher the burden of similarity that the courts will ask plaintiffs to satisfy in order to prove copyright infringement \cite{lemley1995}. For example, in \cite{AppleComputerVMicrosoftCorp}, the court found that Windows interfaces do not infringe Apple’s copyrights because they are similar but not “virtually identical” to the Apple Macintosh unoriginal interfaces. 

Similarly, the less original the copyrighted works, the more likely the courts are to find that the unauthorized uses of these works are fair use \cite{lemley1995}. For example, in \textit{Oracle v. Google} \cite{OracleVGoogle}, the court found that although Google copied verbatim 37 lines of “declaring code” from Oracle’s Java program, the copying of these “generic” aspects is non-infringing \cite{OracleVGoogle, Menell1989}. As part of the fair use analysis, courts may also consider the originality of the allegedly infringing work, not just that of the copyrighted work \cite{lemley1997}. Courts might view highly original derivative works as “transformative,” which weigh heavily in favor of finding for fair use \cite{leval1990, CampbellVAcuffRoseMusicInc}. Alas, as with the lack of definition of copyright originality and the scope-delineating doctrines during the courts’ filtration analysis, the doctrines of substantial similarity and fair use are also notoriously vague \cite{Meurer2013}. As multiple copyright scholars explained over the years, the unavoidable vagueness of copyright scope doctrines is leading to systematic inefficiencies, over-protection of preexisting works, and chilling of expressive speech \cite{gibson2007, litman2008, tehranian2007, samuelson1996, vaidhyanathan2003}

\subsection{Preliminary Technical Background}
In this work, we consider publicly available variants of state-of-the-art generative models. Specifically, we explore Stable Diffusion for image generation (Text-to-Image) and GPT for text generation. To that end we first offer a short technical review of those models.

\paragraph{GPT-2:}
Large language model (LLM) \cite{brown2020language} trained on a massive corpus of text and code, and can generate text, translate languages, write different kinds of creative content, and answer questions in an informative manner. GPT-2 is built upon the transformer architecture and is designed with 1.5 billion parameters. The model is adapted to perform a wide range of natural language processing tasks without task-specific training, including translation, summarization, and content generation \cite{NIPS2017_3f5ee243}.  

GPT-2 is an autoregressive language model. In the autoregressive approach, a sequence of tokens (words or subwords) is generated one token at a time, with each new token being conditioned on the previous ones. Mathematically, the probability of a sequence $s = (t_1,t_2, \cdots ,t_N)$ is factorized as:$P(s) = \prod_{i=1}^{N}P(t_i|t_{<i})$ Where $x_{<i}$ represents all the tokens before the $i^{th}$ token. Each conditional probability $P(t_i|t_{<t})$ is estimated using the transformer architecture, which allows the model to capture long-range dependencies between tokens effectively. 

Assuming GPT-2’s transformer has $L$ layers and a hidden dimension $d$. The model uses an input/output embedding matrix  $E \in \mathbb{R}^{|V| \times d}$, where V represents the vocabulary. For any token in the input sequence and at any layer $l$, the corresponding output is $h^l_i$. The model predicts the token $t_i$ by projecting its final hidden state $h^L_i$ onto the embedding matrix. The final token prediction is then obtained by selecting the token corresponding to the maximum value in this distribution, which can be formulated as: 
\begin{equation}
\hat{t}_l = \mathrm{arg\,max}_{t \in \mathcal{V}} \ Eh_i^L
\end{equation}

By conditioning on prior tokens, GPT-2 can effectively generalize based on the natural language biases it encountered during its training on vast amounts of internet text. This training data, which inherently captures a snapshot of the world's information and perspectives at the time of collection, enables the model to not only reproduce language patterns but also to discern syntactic and semantic similarities. While its performance is commendable, GPT-2 also sparked a discussion about ethical concerns in AI, given its ability to generate coherent yet fabricated content. OpenAI initially withheld the model from public release, citing these concerns, but later provided a staged release as the community became familiarized with its potential and limitations \cite{OpenAI2019BetterLanguageModels}. 

\paragraph{Stable Diffusion}
Stable diffusion is a latent text-to-image diffusion model that was first introduced in the paper "High-Resolution Image Synthesis with Latent Diffusion Models" by  \citet{rombach2022high}. It works by gradually denoising a latent representation of an image, while conditioning the process on a text prompt. Stable diffusion models are trained on a large dataset of image-text pairs and can generate high-quality images from a wide variety of text prompts. The denoising process in stable diffusion can be formally described as follows: Let $x_t$ be the latent representation of the image at time $t$, and let $\mathcal{N}(\cdot)$
 be a Gaussian noise function. The denoising process is then given by the following equation: 
 \begin{equation}
     x_{t+1} = \alpha(x_t) + \mathcal{N}(\sigma)
 \end{equation}
 where $\alpha$ is a denoising function, and $\sigma$ is the standard deviation of the Gaussian noise. The denoising function  $\alpha$ is a neural network that is trained to remove noise from images. The denoising process is repeated iteratively until the desired level of noise reduction is achieved. The denoising process in stable diffusion is conditioned on a text prompt. This is done by passing the text prompt to the denoising function  $\alpha$, implemented using cross-attention mechanisms \cite{NIPS2017_3f5ee243}. This conditioning process allows stable diffusion models to generate images that are consistent with the text prompt. For example, if the text prompt is “a cat sitting on a mat,” the denoising function will generate an image of a cat sitting on a mat. One of the key advantages of stable diffusion models is that they are very stable and easy to control. This makes them well-suited for a variety of applications, such as generating realistic images for creative projects, or creating synthetic images for training other machine learning models.

\subsection{Utilizing Data-Driven Bias in GenAI Models to Evaluate Copyright Genericity}
\begin{figure}[h]
    \setlength{\belowcaptionskip}{-8pt}
    \begin{center}
    \includegraphics[scale=0.9]{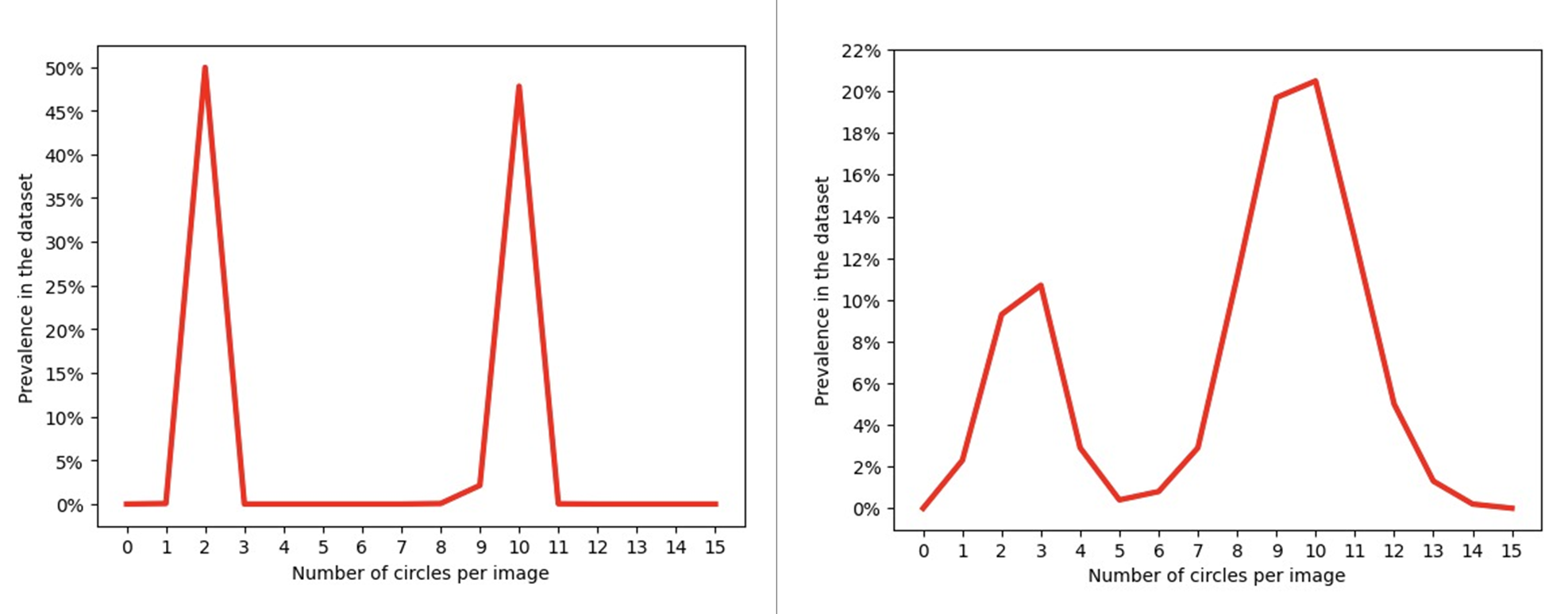}
    \end{center}
    \caption{
    Distributions of circle occurrences in images using a stable diffusion model: (Left) Training dataset, specifically images containing either 2 or 10 circles; (Right) Distribution observed in the generated images.}
    \label{figure:syn}
\end{figure}

By detecting correlations and generalizing patterns from massive corpora of preexisting expressive knowledge, GenAI models can offer a unique opportunity to detect genericity and assess the originality of creative works. As humans, we routinely engage with the corpus of preexisting materials, learning from images, styles, themes, colors, compositions, and the like. Humans memorize impressions, extract principles and generalize from new materials they observe, deconstruct, and reconstruct. All these processes take place exclusively in the silo of the human mind. In an analogy, GenAI also learns from engagement with preexisting materials, but on a much grander scale. Its capacity to learn from data at different levels of granularity reveals some underlying shared patterns in preexisting works, which have been difficult to accurately measure thus far. During learning, models conceptually identify and prioritize expressive elements, as well as rank them according to their frequency in the model’s dataset, which can potentially be harnessed to allow assessing expressive genericity at scale. Generative models tend to produce outputs that reflect the most common patterns found in their training data. If a pattern is frequent (“generic”) in the dataset, the model is more likely to reproduce it. On the other hand, less frequent (“original”) patterns are less commonly produced \cite{brown2020language}. This inherent "bias" stemming from data is a core aspect of machine learning, influencing how GenAI models generalize \cite{liu2022self}. To illustrate this bias, we conduct the following experiment: we train a stable diffusion model from scratch only on images featuring 2 and 10 circles with a white background. The model was trained on a synthetic dataset of size 10K, and we generated 1K images (examples shown in Figure~\ref{figure:syn}). We then compare the distributions of circle occurrences in the training set vs the generated set. As seen in Figure 2. The model was able to generalize and generate images with unseen number of circles in the training set, however it is biased toward generating images with the same number of circles as in the original data. Similar experiments have been performed on past GenAI models such as GANs and VAEs with similar results \cite{10.1145/3528233.3530698}. 

\begin{figure*}[t]
    \setlength{\belowcaptionskip}{-10pt}
    \begin{center}
    \includegraphics[scale=0.5]{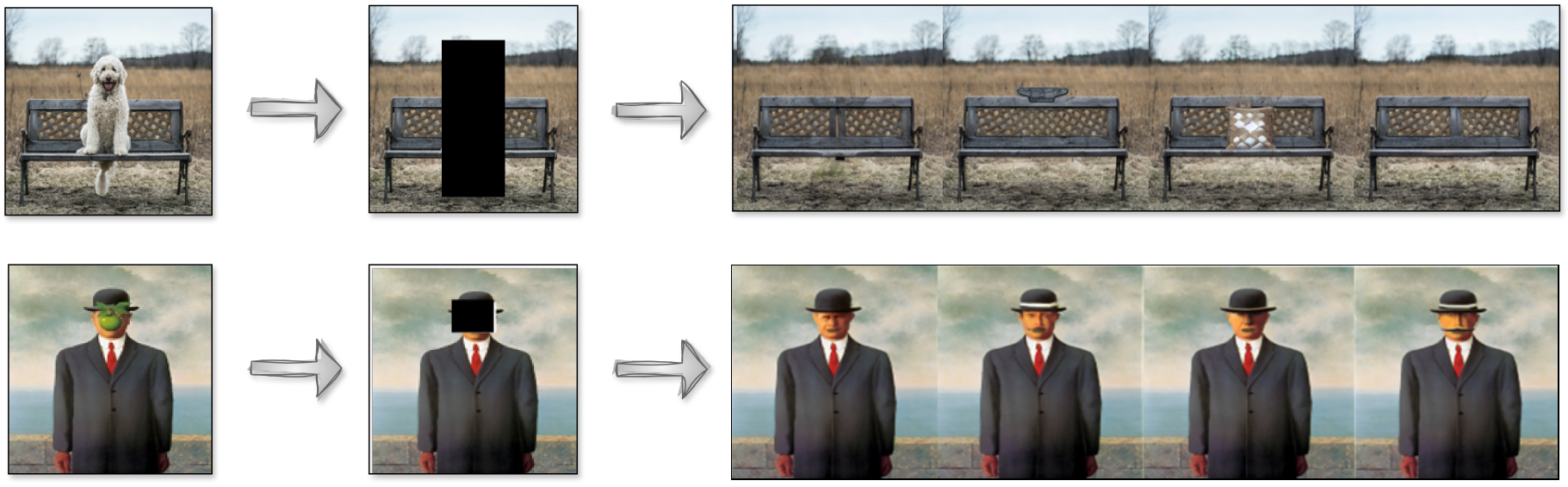}
    \end{center}
    \caption{an inpainting exercise to reconstruct the bench from a photo of a dog on a bench (Top) and apple in René Magritte’s painting The Son of Man (Bottom).}
    \label{fig:inpainting}
\end{figure*}

The same dynamics may be demonstrated using an inpainting technique, which requires GenAI models to reconstruct missing parts from images \cite{rombach2022high}. As shown in Figure~\ref{fig:inpainting}, when we tasked the publicly available pretrained Stable Diffusion with completing a photo of a bench that initially included a dog, the model reconstructed the bench and removed the dog from the scene, as can be expected. This behavior indicates the model has encountered more empty benches in the training set than benches with dogs, and hence is biased towards generating them. Similarly, in the artwork domain, when masking the apple from René Magritte’s famous painting The Son of Man, the model reconstructed the image with human male faces rather than with apples \cite{carlini2021extracting}.\footnote{the reason for this is that the model was trained (unsurprisingly) on many more images of men’s faces rather than men with apples in front of their faces.}

Lastly, in the context of LMs, we display this phenomenon on the publicly available pretrained GPT2 model, by prompting prefix of idioms (all words except for the last word), demonstrating the model's natural bias. For example, when prompting the model with the idiom “put words in someone’s,” it completes it to “mouth” with a high probability of 0.74 as it was likely seen in the training set frequently \cite{haviv2023understanding}. On the other hand, when prompting the model “play it by” instead of predicting “ear”, it predicts another plausible continuation “yourself” with probability of 0.37. These findings are consistent with copyright law. Generic benches and facial human expressions are unprotected by copyright law. The same is true for idioms and other common expressions \cite{CFR2004,PermaGreetings1984}. Thus, the positive correlation between expressive genericity and the GenAI models’ likelihood of recreation is illuminating to copyright law. Since the level of genericity confines the legal protection that the law affords expressive works, 1 We tried this exercise also with adding to the model the textual prompt “Magritte the Son of Man”, but the outputs were still dominated by men’s faces.\footnote{This outcome may also demonstrate that the model’s training data might be biased in favor of men images wearing a suit and a hat, rather than images of women. But it is difficult to tell. Given that the body shape is clearly masculine, and so are other cues in the outfit, the model may properly recognize the appropriate gender.} Courts and policymakers could use this measurement for delineate the scope of copyright protection by measuring the genericity of expressive elements. 

\section{Discussion}
\label{sec:discussion}

The approach proposed by this paper illuminates some limitations in contemporary research. For example, in some cases, researchers may wrongly interpret the literature and equate data memorization in GenAI models with copyright infringement \cite{chang-etal-2023-speak,sag2023copyright,bracha2023work}. For example, \citet{carlini2023extracting}, and \citet{somepalli2023diffusion} have used data-extraction attacks to show that GenAI models can reconstruct specific copyrighted works that appear in the GenAI models’ datasets. However, such a conclusion is inaccurate from a legal standpoint. To the extent that the models’ datasets reflect the cultural environment of a particular expressive domain accurately, memorization is likely to be lawful, even desirable, from a legal perspective \cite{elkin2023can}. When the training models’ datasets are biased towards certain expressive elements, these elements are likely to be considered “generic” and, accordingly, narrowly protected by copyright law. Thus, memorization that is in line with the data-driven bias does not indicate copyright infringement. To ensure that GenAI models sufficiently represent the cultural state of expressive domains, model developers might need to de-duplicate and curate their datasets from pirated or otherwise unlawful material \cite{kandpal2022deduplicating,lee-etal-2022-deduplicating}. They might also need to make sure that the dataset is comprehensive enough and updated periodically to reflect the dynamic evolution of the changing expressive
discourse \cite{lund2009copyright}. Our approach also illuminates why any rigid or binary solutions for ruling out copyright infringement are misguided. For example, one line of research seeks to harness differential privacy or other algorithmic stability approaches to “safeguard” generative models from engaging in copyright infringement \cite{vyas2023provable}. These approaches limit model production to outputs that are not over-influenced by any particular copyrighted work within their data set.
The intuition that guides these approaches-allowing outputs to draw only on common expressive features that are not original to any particular work—is correct. Nevertheless, algorithmic stability approaches cannot serve as a definitive binary test for copyright infringement \cite{henderson2023foundation,elkin2023can}. For example, a model that was trained on many derivatives of an original copyrighted work may produce an infringing derivative that work even if the work itself did not appear in the training
dataset \cite{grimmelmann2023talkin}. The same criticism may be extended to a broader body of research that proposes to employ output filters to prevent copyright infringement \cite{ziegler2021github}. Other than being hackable \cite{ippolito-etal-2023-preventing,henderson2023foundation}, the main problem with filters is that they must boil down flexible legal standards (in our case: substantial similarity of original expression) into rigid rules \cite{kaplow1992rules}. However, as a rich body of scholarship in the context of copyright \cite{bartholomew2014death,boroughf2015next} and elsewhere \cite{dwork2012fairness,kairouz2021advances} has shown, this approach is inadvisable as it may lead to systematic over or under-enforcement. For this reason, in lieu of rigid binary standards, this paper proposes to harness GenAI models to indicate the extent to which an output is already generic, and, therefore, worthy of weaker protection. Using data-driven bias as a metric for (lack of) originality will not provide a definitive answer as to
whether GenAI models’ outputs infringe the copyrights of works used in their training. Nevertheless, genericity measurements would be a valuable input for courts and policymakers, and
inform their legal analysis when grappling with these issues. Specifically, armed with the ability to identify generic patterns in copyrighted works, courts and litigants could better reflect on the sakes of the litigated conflicts, rule in a more predictable manner in tough cases, and avoid unnecessary litigation in easy ones. In addition, the Copyright Office could substitute its current rigid guidelines which denies copyright registration (and protection) for all GenAI works \cite{federalregister2023},with a more nuanced standard that accommodate registration (and protection) for highly original works \cite{miller2009hoisting,parchomovsky2009originality}. 
\section{Conclusions}
This paper has introduced a novel approach that leverages GenAI data-driven biases to measure genericity in copyrighted works on a large scale. Our approach may introduce more nuance to copyright analysis, thereby avoiding over protection of elements that should remain in the public domain to better serve the objectives of copyright law. This approach may inform legal practitioners and copyright stakeholders during infringement litigation, and rights’ registration or licensing. Our approach does not advocate for technological solutionism or full automation of the copyright system; rather, it aims to provide evidence that can inform normative tradeoffs made by social institutions. “Genericity scores,” as proposed in this
article, have the potential to empower policymakers in devising more effective copyright policies and doctrines that align with the evolving landscape of "cheap creativity" facilitated by GenAI. Ultimately, our method can offer valuable insights to policymakers as they navigate the challenges of adapting copyright law to the digital age.

\paragraph{Acknowledgments} 
This research was funded in part by an ERC grant (FOG, 101116258), as well as an ISF Grant (2188 $\backslash$ 20).
Views and opinions expressed are however those of the author(s) only and do not necessarily reflect those of the European Union or the European Research Council Executive Agency. Neither the European Union nor the granting authority can be held responsible for them. In addition, the research leading to these results was supported by TILabs Tel-Aviv University Innovation Labs.

\newpage

\bibliography{anthology,custom}
\bibliographystyle{acl_natbib}

\appendix

\end{document}